\documentclass[11pt,a4paper]{article}
\usepackage[hyperref]{acl2017}
\usepackage{times}
\usepackage{latexsym}
\usepackage{hyperref}
\usepackage{amsmath}
\usepackage{color}
\usepackage{amsfonts}
\usepackage{amssymb}
\usepackage{graphbox}
\usepackage{enumitem}
\usepackage[utf8]{inputenc}

\usepackage{url}
\newcommand{\denselist}{\setlength{\itemsep}{1pt}
  \setlength{\parskip}{0pt} \setlength{\parsep}{0pt}}
\newcommand{\bitem}{\begin{itemize}[noitemsep,topsep=2pt]\denselist}
\newcommand{\eitem}{\end{itemize}}

\newcommand{\bx}{\mathbf{x}}
\newcommand{\by}{\mathbf{y}}

\newcommand{\bh}{\mathbf{h}}

\newcommand{\bw}{\mathbf{w}}
\newcommand{\bm}{\mathbf{m}}
\newcommand{\br}{\mathbf{r}}

\newcommand{\eat}[1]{\ignorespaces}
\newcommand{\commentout}[1]{}

\newcommand{\starts}{s^{\dagger}}

\newcommand{\ignore}[1]{}

\usepackage{booktabs}

\DeclareMathOperator*{\argmax}{argmax}

\aclfinalcopy 

\title{DRAGNN: A Transition-based Framework for Dynamically Connected Neural Networks}

\author{Lingpeng Kong$^\dagger$ \;\; Chris Alberti$^\star$ \;\; Daniel Andor$^\star$ \;\; Ivan Bogatyy$^\star$ \;\; David Weiss$^\star$\\
  {\tt \small lingpengk@cs.cmu.edu, \{chrisalberti,danielandor,bogatyy,djweiss\}@google.com} \\
  $^\dagger$Carnegie Mellon University, Pittsburgh, PA. \\
  $^\star$Google, New York, NY.
}

\date{}

\begin{document}
\maketitle


\begin{abstract}
  In this work, we present a compact, modular framework for constructing novel
  recurrent neural architectures. Our basic module is a new generic unit, the
  Transition Based Recurrent Unit (TBRU). In addition to hidden layer
  activations, TBRUs have discrete state dynamics that allow network
  connections to be built dynamically as a function of intermediate
  activations. By connecting multiple TBRUs, we can extend and combine commonly
  used architectures such as sequence-to-sequence, attention mechanisms, and
  recursive tree-structured models. A TBRU can also serve as both an {\em
    encoder} for downstream tasks and as a {\em decoder} for its own task
  simultaneously, resulting in more accurate multi-task learning. We call our
  approach Dynamic Recurrent Acyclic Graphical Neural Networks, or DRAGNN. We
  show that DRAGNN is significantly more accurate and efficient than seq2seq
  with attention for syntactic dependency parsing and yields more accurate
  multi-task learning for extractive summarization tasks.
\end{abstract}

\section{Introduction}

\begin{figure*}[t]
  \centering
  \includegraphics[width=1.0\linewidth]{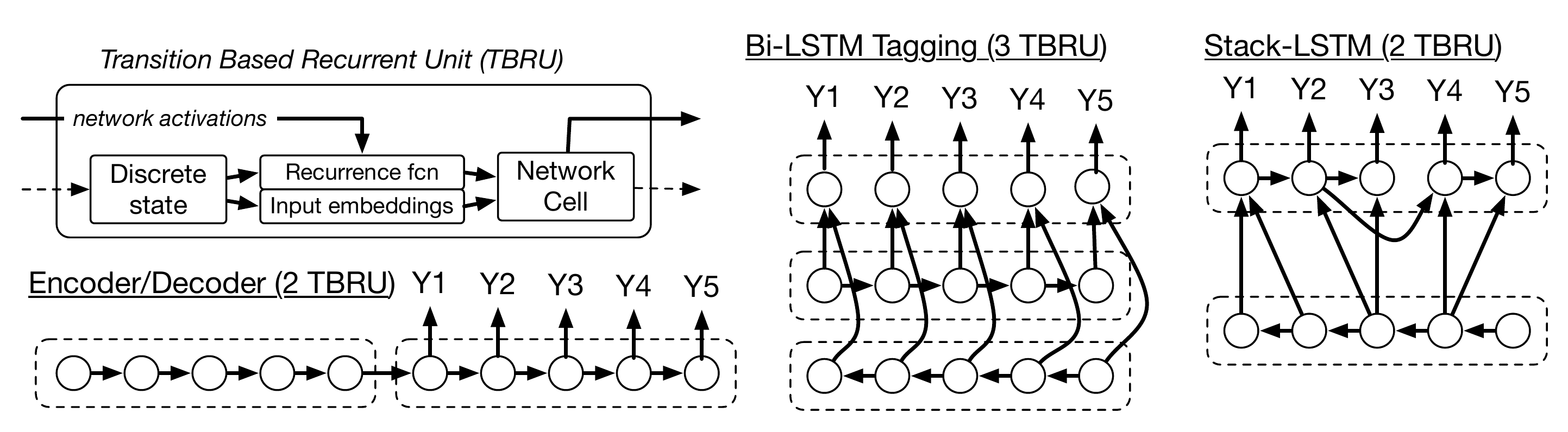}
  \caption{High level schematic of a Transition-Based Recurrent Unit (TBRU), and
    common network architectures that can be implemented with multiple TBRUs.
    The discrete state is used to compute recurrences and fixed input
    embeddings, which are then fed through a network cell. The network predicts
    an action which is used to update the discrete state (dashed output) and
    provides activations that can be consumed through recurrences (solid
    output). Note that we present a slightly simplified version of Stack-LSTM
    \cite{dyer2015transition} for clarity.}
  \label{fig:overview}
\end{figure*}

\begin{figure*}[t]
  \centering
  \includegraphics[width=1.0\linewidth]{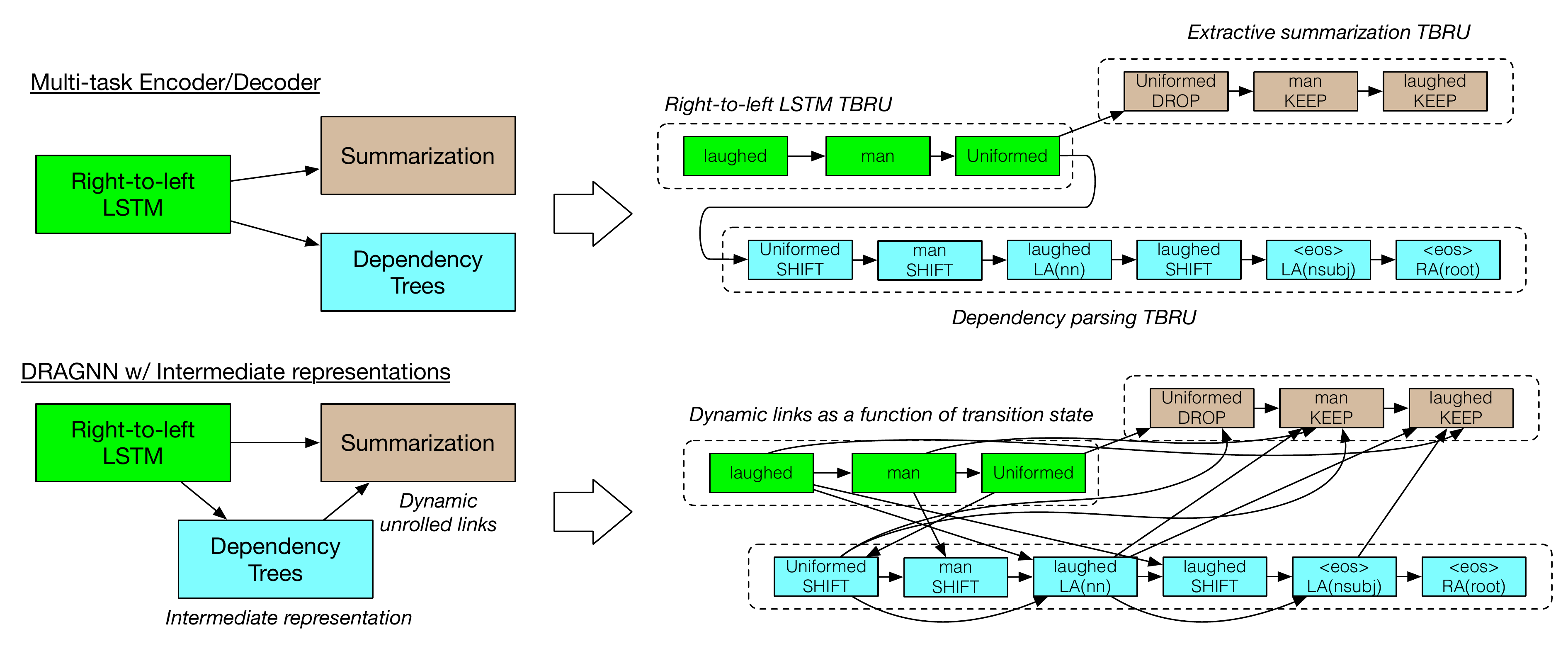}
  \caption{Using TBRUs to share fine-grained, structured representations. Top
    left: A high level view of multi-task learning with DRAGNN in the style of
    multi-task seq2seq \cite{DBLP:journals/corr/LuongLSVK15}. Bottom
    left: Extending the ``stack-propagation'' \cite{zhang2016stack} idea to
    included dependency parse trees as intermediate representations. Right:
    Unrolled TBRUs for each setup for a input fragment ``Uniformed man
    laughed'', utilizing the transition systems described in Section
    \ref{sec:experiments}.
  }
  \label{fig:overview-multi-task}
\end{figure*}

To apply deep learning models to structured prediction, machine learning
practitioners must address two primary issues: (1) how to represent the input,
and (2) how to represent the output. The {\em seq2seq} encoder/decoder framework
\cite{kalchbrenner13,cho2014learning,sutskever2014sequence} proposes solving
these generically. In its simplest form, the {\em encoder} network produces a
fixed-length vector representation of an input, while the {\em decoder} network
produces a linearization of the target output structure as a sequence of output
symbols. Encoder/decoder is state of the art for several key tasks in natural
language processing, such as machine translation \cite{wu2016google}.

However, fixed-size encodings become less competitive when the input structure can
be explicitly mapped to the output. In the simple case of predicting tags for
individual tokens in a sentence, state-of-the-art taggers learn vector
representations for each input {\em token} and predict output tags from those
\cite{ling2015finding,huang2015bidirectional,andor2016globally}. When the input
or output is a syntactic parse tree, networks that explicitly operate over the
compositional structure of the network typically outperform generic
representations \cite{dyer2015transition,li2015tree, bowman2016fast}. Implictly
learned mappings via attention mechanisms can significantly improve the
performance of sequence-to-sequence \cite{bahdanau15,vinyals2015grammar}, but
require runtime that's quadratic in the input size.

In this work, we propose a modular neural architecture that generalizes the
encoder/decoder concept to include explicit structure. Our framework can
represent sequence-to-sequence learning as well as models with explicit
structure like bi-directional tagging models and compositional, tree-structured
models.
Our core idea is to define any given architecture as a series of modular units,
where connections between modules are unfolded {\em dynamically} as a function
of the intermediate activations produced by the network. These dynamic
connections represent the explicit input and output structure produced by the
network for a given task.

We build on the idea of {\em transition systems} from the parsing literature
\cite{nivre2006inductive}, which linearize structured outputs as a sequence of
({\em state}, {\em decision}) pairs. Transition-based neural networks have
recently been applied to a wide variety of NLP problems;
\newcite{dyer2015transition,lample2016neural,kiperwasser2016simple,zhang2016transition,andor2016globally},
among others. We generalize these approaches with a new basic module, the
Transition-Based Recurrent Unit (TBRU), which produces a vector representation
for every transition state in the output linearization (Figure
\ref{fig:overview}). These representations also serve as the encoding of the
explicit structure defined by the states. For example, a TBRU that attaches two
sub-trees while building a syntactic parse tree will also produce the hidden
layer activations to serve as an encoding for the newly constructed
phrase. Multiple TBRUs can be connected and learned jointly to add explicit
structure to multi-task learning setups and share representations between tasks
with different input or output spaces (Figure \ref{fig:overview-multi-task}).

This inference procedure will construct an acyclic compute graph representing
the network architecture, where recurrent connections are dynamically added as
the network unfolds. We therefore call our approach Dynamic Recurrent Acyclic
Graphical Neural Networks, or DRAGNN.

DRAGNN has several distinct modeling advantages over traditional fixed neural
architectures. Unlike generic seq2seq, DRAGNN supports variable sized input
representations that may contain explicit structure. Unlike purely sequential
RNNs, the dynamic connections in a DRAGNN can span arbitrary distances in the
input space. Crucially, inference remains linear in the size of the input,
in contrast to quadratic-time attention mechanisms.
Dynamic connections thus establish a compromise between pure seq2seq
and pure attention architectures by providing a finite set of long-range 
inputs that `attend' to relevant portions of the input space.
Unlike recursive neural networks \cite{socher2010learning,socher2011dynamic}
DRAGNN can both predict intermediate structures (such as parse trees) and utilize
those structures in a single deep model, backpropagating downstream task errors
through the intermediate structures.
Compared to models such as Stack-LSTM \cite{dyer2015transition}
and SPINN \cite{bowman2016fast}, TBRUs are a more general formulation that
allows incorporating dynamically structured multi-task learning
\cite{zhang2016stack} and more varied network architectures.

In sum, DRAGNN is not a particular neural architecture, but rather a {\em
  formulation for describing neural architectures compactly.} The key to this
compact description is a new recurrent unit---the TBRU---which allows connections
between nodes in an unrolled compute graph to be specified dynamically in a generic
fashion. We utilize transition systems to provide succinct, discrete
representations via linearizations of both the input and the output for
structured prediction. We provide a straightforward way of
re-using representations across NLP tasks that operate on different structures.

We demonstrate the effectiveness of DRAGNN on two NLP tasks that benefit from
explicit structure: dependency parsing and extractive sentence summarization
\cite{filippova2013overcoming}. First, we show how to use TBRUs to
incrementally add structure to the input and output of a ``vanilla'' seq2seq
dependency parsing model, dramatically boosting accuracy over seq2seq with no
additional computational cost. Second, we demonstrate how the same TBRUs can be used to
provide structured intermediate syntactic representations for extractive
sentence summarization. This yields better accuracy than is possible with the
generic multi-task seq2seq \cite{dong2015multi,DBLP:journals/corr/LuongLSVK15}
approach. Finally, we show how multiple TBRUs for
the same dependency parsing task can be stacked together to produce a single
state-of-the-art dependency parsing model.


\begin{figure*}[t]
  \centering
  \includegraphics[width=0.48\linewidth]{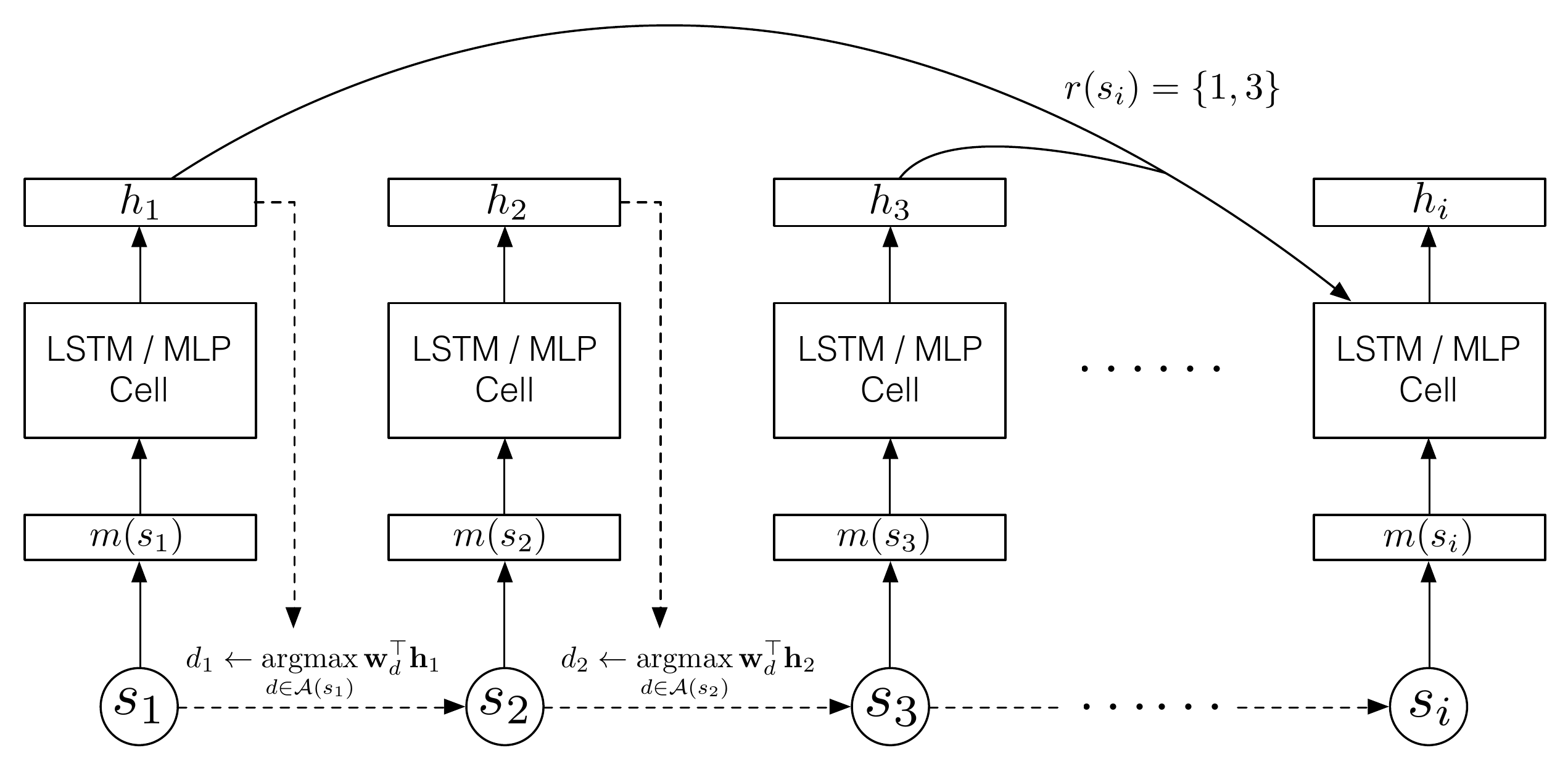}
  \includegraphics[width=0.48\linewidth]{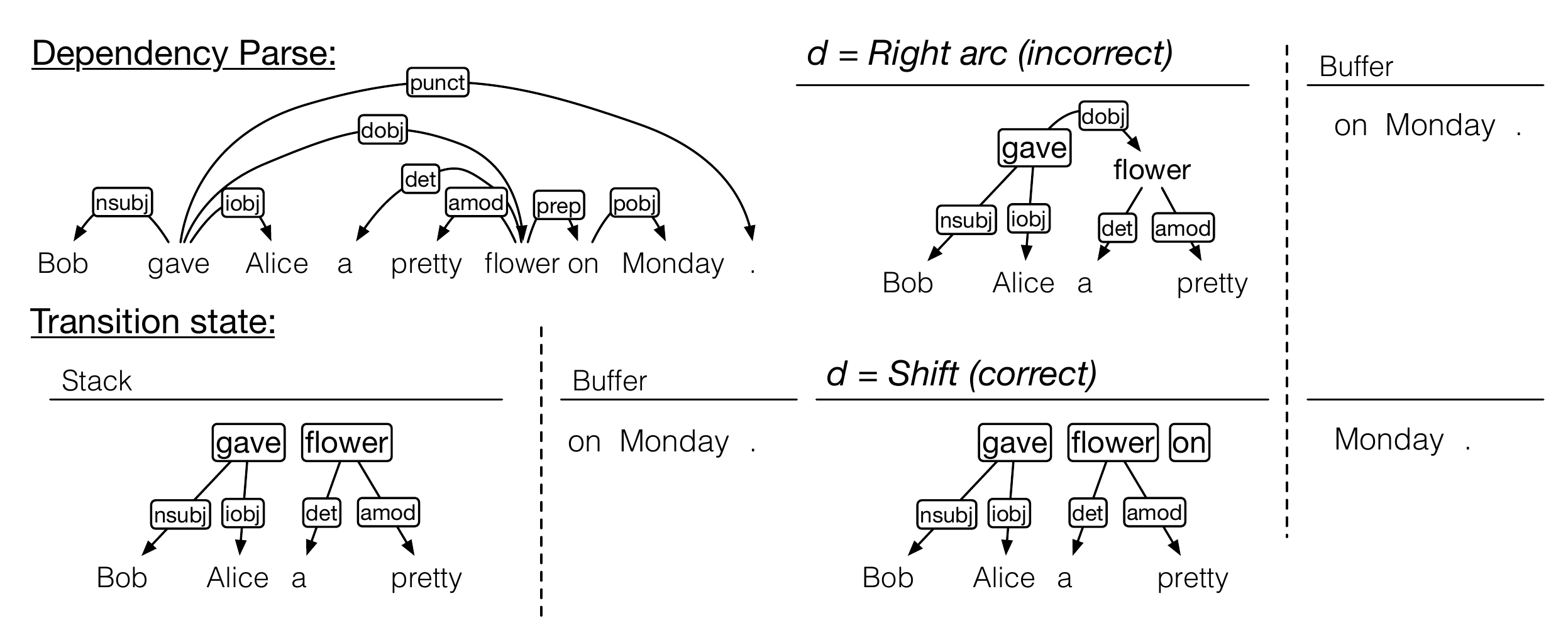}
  \caption{Left: TBRU schematic. Right: Dependency parsing example. A gold 
    parse tree and an {\em arc-standard} transition state
    with two sub-trees on the stack are shown. From this state, two possible
    actions are also shown ({\em Shift} and {\em Right arc}). 
	To agree with the gold tree, the {\em Shift} action should be taken.}
  \label{fig:tbru}
\end{figure*}

\section{Transition Systems}
\label{sec:model}




We use {\em transition systems} to map inputs $x$ into a sequence of output
symbols, $d_1 \ldots d_n$. For the purposes of implementing DRAGNN, transition systems make
explicit two desirable properties. First, we stipulate that the output symbols
represent modifications of a persistent, discrete {\em state}, which makes
book-keeping to construct the dynamic recurrent connections easier to
express. Second, transition systems make it easy to enforce arbitrary
constraints on the output, e.g. the output should produce a valid tree.

Formally, we use the same setup as \newcite{andor2016globally}, and define a
transition system $\mathcal{T} = \{\mathcal{S}, \mathcal{A}, t\}$ as:
\begin{itemize}
\item A set of states $\mathcal S(x)$.
\item A special start state $\starts \in\mathcal S(x)$.
\item A set of {\em allowed} decisions ${\cal A}(s, x)$ for all $s\in\mathcal S$.
\item A transition function $t(s, d, x)$ returning a new state $s'$ for
  any decision $d\in {\cal A}(s, x)$.
\end{itemize}
For brevity, we will drop the dependence on $x$ in the functions given
above. Throughout this work we will use transition systems in which all complete
structures for the same input $x$ have the same number of decisions $n(x)$ (or
$n$ for brevity), although this is not necessary.

A complete structure is then a sequence of decision/state pairs
$(s_1, d_1) \ldots (s_n,d_n)$ such that $s_1 = \starts$, $d_i \in {\cal A}(s_i)$
for $i = 1 \ldots n$, and $s_{i+1} = t(s_i, d_i)$. We will now define recurrent
network architectures that operate over these linearizations of input and output
structure.

\section{Transition Based Recurrent Networks}

We now formally define how to combine transition systems with recurrent
networks into what we call a {\em transition based recurrent unit} (TBRU). A
TBRU consists of the following:

\bitem
\item A transition system $\mathcal{T}$,
\item An input function $\bm(s)$ that maps states to fixed-size vector representations, for example,
an embedding lookup operation for features from the discrete state, $\bm(s): \mathcal{S} \mapsto \mathcal{R}^{K}$
\item A recurrence function $\br(s)$ that maps states to a set of previous time
  steps:
$$\quad \br(s): \mathcal{S}
\mapsto \mathbb{P}\{1, \dots, {i-1}\},$$ where $\mathbb{P}$ is the power
set. Note that in general $|\br(s)|$ is not necessarily fixed and can vary with
$s$. We use $\br$ to specify state-dependent recurrent links in the unrolled
computation graph. 
\item A RNN cell that computes a new hidden representation from the fixed and recurrent inputs:
$$\bh_{s} \leftarrow \mathbf{RNN}(\bm(s),\{\bh_i \mid i \in \br(s)\}).$$
\eitem

\paragraph{Example 1. Sequential tagging RNN.} Let the input $x = \{\bx_1,
\dots, \bx_n\}$ be a sequence of word embeddings, and the output be a sequence of
tags $d_1, \dots, d_n$. Then we can model a simple LSTM tagger as follows:

\bitem
\item $\mathcal{T}$ sequentially tags each input token, where $s_i =
  \{1,\dots,d_{i-1}\}$, and $\mathcal{A}$ is the set of possible tags. We call
  this the {\em tagger} transition system.
\item $\bm(s_i) = \bx_i$, the word embedding for the next token to be tagged.
\item $\br(s_i) = \{i-1\}$ to connect the network to the previous state.
\item $\mathbf{RNN}$ is a single instance of the LSTM cell.
\eitem

\paragraph{Example 2. Parsey McParseface.} The open-source syntactic parsing
model of \newcite{andor2016globally} can be defined in our framework as follows:
\bitem
\item $\mathcal{T}$ is the {\em arc-standard} transition system (Figure \ref{fig:tbru}),
  so the state contains all words and partially built trees on the stack as well
  as unseen words on the buffer.
\item $\bm(s_i)$ is the concatenation of 52 feature embeddings extracted from
  tokens based on their positions in the stack and the buffer.
\item $\br(s_i) = \{\}$ is empty, as this is a feed-forward network.
\item $\mathbf{RNN}$ is a feed-forward multi-layer perceptron (MLP).
\eitem

\paragraph{Inference with TBRUs.} Given the above, inference in the TBRU
proceeds as follows:
\begin{enumerate}[noitemsep]\denselist
\item Initialize $s_1 = \starts$.
\item For $i = 1, \dots, n$:
  \begin{enumerate}
  \item Update the hidden state:\\
    $\bh_{i} \leftarrow \mathbf{RNN}(\bm(s_i),\{\bh_j \mid j \in \br(s_i)\}).$
  \item Update the transition state:\\
    $d_i \leftarrow \argmax_{d \in \mathcal{A}(s_i)}\bw_d^\top \bh_i, \quad s_{i+1} \leftarrow t(s_i, d_i).$
  \end{enumerate}
\end{enumerate}

A schematic overview of a single TBRU is presented in Figure \ref{fig:tbru}. By
adjusting $\mathbf{RNN}$, $\br$, and $\mathcal{T}$, TBRUs can represent a wide
variety of neural architectures.

\begin{figure*}[t]
  \centering
  \includegraphics[width=1.0\linewidth]{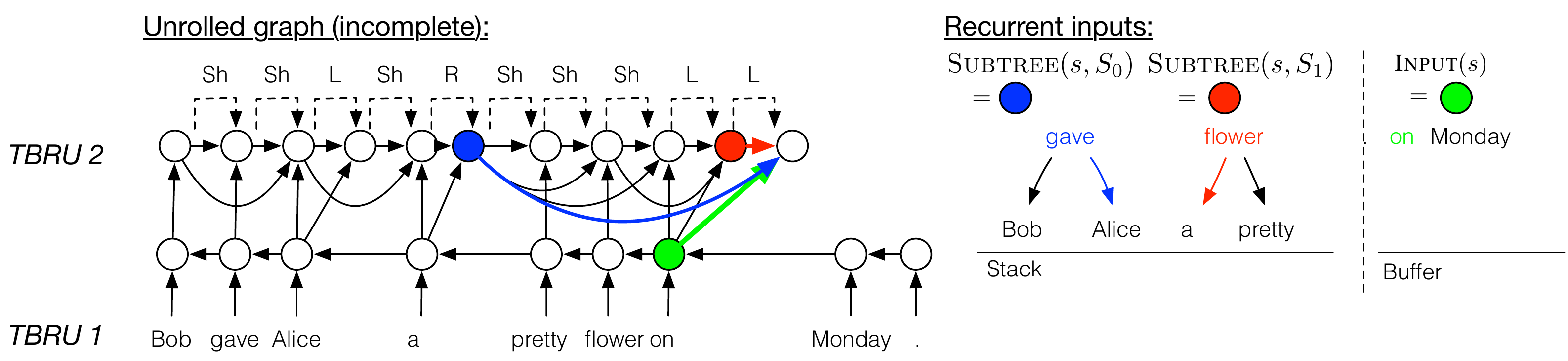}
  \caption{Detailed schematic for the compositional dependency parser used in
    our experiments. {\em TBRU 1} consumes each input word right-to-left. 
    {\em TBRU 2} uses the {\em arc-standard} transition system.
    Note how each {\em Shift} action causes the {\em TBRU 1}$\rightarrow${\em TBRU 2} link to
    advance. The dynamic recurrent inputs to each state are highlighted;
    the stack representations are obtained from the last {\em Reduce} action to
    modify each sub-tree.}
  \label{fig:seq2seq}
\end{figure*}

\subsection{Connecting multiple TBRUs to learn shared representations}
\label{sec:connecting}

While TBRUs are a useful abstraction for describing recurrent models, the
primary motivation for this framework is to allow new architectures by combining
representations across tasks and compositional structures. We do this by
connecting multiple TBRUs with different transition systems via the recurrence
function $\br(s)$. We formally augment the above definition as follows:
\begin{enumerate}[noitemsep]\denselist
\item We execute a list of $T$ TBRU components, one at a time, so that each TBRU
  advances a global step counter. Note that for simplicity, we assume an earlier
  TBRU finishes all of its steps before the next one starts execution.
\item Each transition state from the $\tau$'th component $s^\tau$ has access to
  the terminal states from every prior transition system, and the recurrence
  function $\br(s^\tau)$ for any given component can pull hidden activations
  from every prior one as well.
\end{enumerate}

\paragraph{Example 3. ``Input'' transducer TBRUs via no-op decisions.} We find
it useful to define TBRUs even when the transition system decisions don't
correspond to any output. These TBRUs, which we call {\em no-op} TBRUs,
transduce the input according to some linearization. The simplest is the {\em
  shift-only} transition system, in which the state is just an input pointer
$s_i = \{i\}$, and there is only one transition which advances it: $t(s_i,
\cdot) = \{i+1\}$. Executing this transition system will produce a hidden
representation $\bh_i$ for every input token.

\paragraph{Example 4. Encoder/decoder networks with TBRUs.} We can reproduce the
encoder/decoder framework for sequence tagging by using two TBRUs: one using the
{\em shift-only} transition system to encode the input, and the other using the
{\em tagger} transition system. For input $x = \{\bx_1, \dots, \bx_n\}$, we
connect them as follows:
\bitem
\item For {\em shift-only} TBRU: $\bm(s_i) = \bx_i$, $\br(s_i) =\{i-1\}$.
\item For {\em tagger} TBRU: $\bm(s_{n+i}) = \by_{d_{n+i-1}}$, $\br(s_i) =\{n,
  n+i-1\}$.
\eitem

We observe that the {\em tagger} TBRU starts at step $n$ after the {\em
  shift-only} TBRU finishes, that $\by_j$ is a fixed embedding vector for the
output tag $j$, and that the {\em tagger} TBRU has access to both the final
encoding vector $\bh_n$ as well as its own previous time step $\bh_{n+i-1}$.

\paragraph{Example 4. Bi-directional LSTM tagger.} With three TBRUs, we can
implement a simple bi-directional tagger. The first two run the {\em shift-only}
transition system, but in opposite directions. The final TBRU runs the {\em
  tagger} transition system and concatenates the two representations:
\bitem
\item Left to right: $\mathcal{T} = $ {\em shift-only}, $\bm(s_i) = \bx_i$, $\br(s_i)
  =\{i-1\}$.
\item Right to left: $\mathcal{T} = $ {\em shift-only}, $\bm(s_{n+i}) = \bx_{n-i}$, $\br(s_{n+i})
  =\{n+i-1\}$.
\item Tagger: $\mathcal{T} = tagger$, $\bm(s_{2n+i}) = \{\}$, $\br(s_{2n+i}) =\{i, 2n-i\}$.
\eitem

We observe that the network cell in the tagger TBRU takes recurrences only from
the bi-directional representations, and so is not recurrent in the traditional
sense. See Fig.~\ref{fig:overview} for an unrolled example.

\begin{table*}[t]
  \centering
\scalebox{0.9}{
  \begin{tabular}{ccccc}
    \toprule
    \multicolumn{2}{c}{Parsing TBRU recurrence, $\br(s_i) \subseteq \{1, \dots, n+i\}$} & \multicolumn{2}{c}{Parsing Accuracy (\%)} \\
    \cmidrule(r){1-2}
    Input links & Recurrent edges & News & Questions & Runtime \\
    \midrule
    $\{n\}$ & $\{n+i-1\}$ & 27.3 & 70.1 & $O(n)$ \\
    $\{n\}$ & $\{\textsc{Subtree}(s_i, S_0), \textsc{Subtree}(s_i, S_1)\}$ & 36.0 & 75.6 & $O(n)$\\

    \addlinespace[0.5em]
    Attention & $\{n+i-1\}$ & 76.1 & 84.8 & $O(n^2)$\\
    Attention & $\{\textsc{Subtree}(s_i, S_0), \textsc{Subtree}(s_i, S_1)\}$ & 89.0 & 91.9 & $O(n^2)$\\

    \addlinespace[0.5em]
    $\textsc{Input}(s_i)$ & $\{n+i-1\}$ & 87.1 & 89.7 & $O(n)$\\
    $\textsc{Input}(s_i)$ & $\{\textsc{Subtree}(s_i, S_0), \textsc{Subtree}(s_i, S_1)\}$ & {\bf 90.9} & {\bf 92.1} & $O(n)$\\
    \bottomrule
  \end{tabular}
}
  \caption{Dynamic links enable much more accurate, efficient linear-time parsing models on the Treebank Union dev set. We vary the recurrences $\br$ to explore utilizing explicit structure in the parsing TBRU. Utilizing the explicit $\textsc{Input}(s_i)$ pointer is more effective and more efficient than a quadratic attention mechanism. Incorporating the explicit stack structure via recurrent links further improves performance.\vspace{-1em}}
  \label{tab:parsing}
\end{table*}

\paragraph{Example 5. Multi-task bi-directional tagging.} Here we observe that
it's possible to add additional annotation tasks to the bi-directional TBRU
stack from Example 4 simply by adding more instances of the tagger TBRUs that
produce outputs from different tag sets, e.g. parts-of-speech vs. morphological
tags. Most important, however, is that any additional TBRUs have access to {\em
  all three} earlier TBRUs. This means that we can support the
``stack-propagation'' \cite{zhang2016stack} style of multi-task learning simply
by changing $\br$ for the last TBRU:
\bitem
\item Traditional multi-task:\begin{align*}\br(s_{3n+i}) = \{i, 2n-i\}\end{align*}
\item Stack-prop:\begin{align*}\br(s_{3n+i}) = \{\underbrace{i}_{\text{Left-to-right}}, \underbrace{2n-i}_{\text{Right-to-left}}, \underbrace{2n+i}_{\text{Tagger TBRU}}\}\end{align*}
\eitem

\paragraph{Remark: the {\em raison d'être} of DRAGNN.} This example highlights the
primary advantage of our formulation: {\em a TBRU can serve as both an encoder
  for downstream tasks and as a decoder for its own task simultaneously.} This
idea will prove particularly powerful when we consider syntactic parsing, which
involves compositional structure over the input. For example, consider a no-op
TBRU that traverses an input sequence $\bx_1,\dots,\bx_n$ in the order
determined by a binary parse tree: this transducer can implement a recursive
tree-structured network in the style of \newcite{tai2015improved}, which computes
representations for sub-phrases in the tree. 
In contrast, with DRAGNN, we can use the {\em arc-standard} parser directly to
produce the parse tree as well as encode sub-phrases into representations.

\paragraph{Example 6. Compositional representations from {\em arc-standard}
  dependency parsing.}
We use the {\em arc-standard} transition system \cite{nivre2006inductive} to
model dependency trees.
The system maintains two data structures as part of the state $s$: an input
pointer and a stack (Figure \ref{fig:tbru}). Trees are built bottom up via three
possible attachment decisions. Assume that the stack consists of $S = \{A,B\}$, with the
next token being $C$. We use $S_0$ and $S_1$ to refer to the top two tokens on
the stack. Then the decisions are defined as: \bitem
\item Shift: Push the next token on to the stack: $S = \{A,B,C\}$, and advance the
  input pointer.
\item Left arc + {\em label}: Add an arc $A \leftarrow_{label} B$, and remove $A$ from the stack: $S = \{B\}$.
\item Right arc + {\em label}: Add an arc $A \rightarrow_{label} B$, and remove $B$ from the stack: $S = \{A\}$.
\eitem
For a given parser state $s_i$, we compute two types of recurrences:
\bitem
\item $\br_\textsc{input}(s_i) = \{\textsc{Input}(s_i)\}$, where \textsc{Input}
  returns the index of the next input token.
\item $\br_\textsc{stack}(s_i)$ =\\
  $ \quad\quad\{\textsc{Subtree}(s_i, S_0), \textsc{Subtree}(s,
  S_1)\}$,\\
  where $\textsc{Subtree(s,i)}$ is a function returning the index of
  the last decision that modified the $i$'th token:
\begin{align*}
&\textsc{Subtree}(s,i) \\
&\quad = \argmax_j \{ d_j \textrm{ s.t. } d_j \textrm{ shifts or adds a new} \\
  &\quad\quad \quad\quad\textrm{child to token } i \}
\end{align*}

\eitem We show an example of the links constructed by these recurrences in
Figure \ref{fig:seq2seq}, and we investigate variants of this model in Section
\ref{sec:experiments}. This model is recursively compositional according to the
decision taken by the network: when the TBRU at step $s_i$ decides to add an arc
$A\rightarrow B$ for state, the activations $\bh_i$ will be used to represent
that new subtree in future decisions.\footnote{This composition function is
  similar to that in the constituent parsing SPINN model \cite{bowman2016fast},
  but with several key differences. Since we use TBRUs, we compose new
  representations for ``Shift'' actions as well as reductions, we take
  inputs from other recurrent models, and we can utilize subtree representations
  in downstream tasks.}

\paragraph{Example 7. Extractive summarization pipeline with parse
  representations.}
To model extractive summarization, we follow \newcite{andor2016globally} and use a
{\em tagger} transition system with two tags: {\em Keep} and {\em Drop}. However,
whereas \newcite{andor2016globally} use discrete features of the parse tree, we can
utilize the $\textsc{subtree}$ recurrence function to pull compositional,
phrase-based representations of tokens as constructed by the dependency
parser. This model is outlined in Fig.~\ref{fig:overview-multi-task}. 
A full specification is given in the Appendix.





\begin{table*}[t]
  \centering
\scalebox{0.9}{
  \begin{tabular}{ccccc}
    \toprule
    Input representation & Multi-task style & A (\%) & F1 (\%) & LAS (\%) \\
    \midrule
    Single LSTM & -- & 28.93 & 79.75 & -- \\
    Bi-LSTM  & -- & 29.51 & 80.03 & --\\
    Multi-task LSTM & \newcite{DBLP:journals/corr/LuongLSVK15} & 30.07	& 80.31 & 89.42 \\
    Parse sub-trees (Figure \ref{fig:overview-multi-task}) & \newcite{zhang2016stack} & {\bf 30.56} & {\bf 80.74} & 89.13 \\
    \bottomrule
  \end{tabular}
}
\caption{Single- vs.\ multi-task learning with DRAGNN on extractive summarization. ``A'' is full-sentence accuracy of the extraction model, ``F1'' is per-token F1 score, and ``LAS'' is labeled parsing accuracy on the Treebank Union News dev set. Both multi-task models that use the parsing data outperform the single-task approach, but the model that uses parses as an intermediate representation via our extension of \newcite{zhang2016stack} (Fig.~\ref{fig:overview-multi-task}) is more effective.
  The locally normalized model in \newcite{andor2016globally} obtains 30.50\% accuracy and 78.72\% F1.
}
  \label{tab:compress}
\end{table*}

\subsection{How to train a DRAGNN}
\label{sec:train}

Given a list of TBRUs, we propose the following learning
procedure. We assume training data consists of examples $x$ along with gold
decision sequences for one of the TBRUs in the
DRAGNN. At a minimum, we need such data for the final TBRU.
Given decisions $d_1 \dots d_N$ from prior components $1 \dots T-1$, we
define a log-likelihood objective to train the $T$'th TBRU along its gold
decision sequence $d_{N+1}^\star, \dots, d_{N+n}^\star$:
\begin{align}
\label{eq:obj}
  &L(x,d^{\star}_{N+1:N+n};\theta) = \nonumber\\
  &\quad\sum_{i} \log P(d_{N+i}^\star \mid d_{1:N}, d^\star_{N+1:N+i-1}; \theta) ,
\end{align}
where $\theta$ are the combined parameters across all TBRUs.
Eq.~\eqref{eq:obj} is locally normalized \cite{andor2016globally},
since we optimize the probabilities of the individual decisions in the gold
sequence.

The remaining question is where the decisions $d_1 \dots d_N$ come from.
There are two options here: either 1) they come as part of the gold annotation
(e.g.\ if we have joint tagging and parsing data), or 2) they are predicted by
unrolling the previous components. When training the stacked extractive
summarization model, the parse trees will be predicted by the previously
trained parser TBRU.

When training a given TBRU, we unroll an entire input sequence and then use
backpropagation through structure \cite{goller1996learning} to optimize
\eqref{eq:obj}. To train the whole system on a set of $C$ datasets, we use a
strategy similar to \cite{dong2015multi,DBLP:journals/corr/LuongLSVK15}. We
sample a target task $c, 1 \leq c \leq C$, based on a pre-defined distribution, and take a
stochastic optimization step on the objective of task $c$'s TBRU. In practice,
task sampling is usually preceded by a deterministic number of pre-training
steps, allowing, for example, to run a certain number of tagger training
steps before running any parser training steps.

\newcommand{\ra}{\rightarrow}
\begin{table*}[t]
  \centering
  \begin{tabular}{cccccccccc}
  \toprule
    & \multicolumn{3}{c}{Union-News} &\multicolumn{3}{c}{Union-Web} &\multicolumn{3}{c}{Union-QTB} \\
    Model & UAS & LAS & POS & UAS & LAS & POS & UAS & LAS & POS \\ 
    \midrule
    Andor et al. (2016) & 94.44 &92.93 &97.77 &90.17 &87.54 &94.80 &95.40 &93.64 & 96.86 \\
    Left-to-right parsing & 94.60 & 93.17 & 97.88& 90.09 & 87.50 &94.75 & 95.62 &94.06 & 96.76\\
    Deep stacked parsing & {\bf 94.66} & {\bf 93.23} & {\bf 98.09} & {\bf 90.22} & {\bf 87.67} & {\bf 95.06} & {\bf 96.05} & {\bf 94.51} & {\bf 97.25}  \\
    \bottomrule
  \end{tabular}
  \caption{Deep stacked parsing compared to state-of-the-art on Treebank Union for parsing and POS.}
  \label{tab:deep-bidir}
\end{table*}

\section{Experiments}
\label{sec:experiments}

In this section, we evaluate three aspects of our approach on two NLP tasks:
English dependency parsing and extractive sentence summarization. For English
dependency parsing, we primarily use the Union Treebank setup from
\newcite{andor2016globally}. By evaluating on both news and questions domains, we
can separately evaluate how the model handles naturally longer and shorter form
text.
On the Union Treebank setup there are 93 possible actions considering all
arc-label combinations. For extractive sentence summarization, we use the
dataset of \newcite{filippova2013overcoming}, where a large news collection is used
to heuristically generate compression instances. The final corpus contains about
2.3M compression instances, but since we evaluated multiple tasks using this
data, we sub-sampled the training set to be comparably sized to the parsing data
($\approx$60K training sentences). The test set contains 160K examples.
We implement our method in TensorFlow, using mini-batches of size 4 and
following the averaged momentum training and hyperparameter tuning procedure of
\newcite{weiss2015structured}.

\paragraph{Using explicit structure improves encoder/decoder}

We explore the impact of different types of recurrences on dependency parsing in
Table \ref{tab:parsing}. In this setup, we used relatively small models:
single-layer LSTMs with 256 hidden units, taking 32-dimensional word or output
symbol embeddings as input to each cell. In each case, the parsing TBRU takes
input from a right-to-left {\em shift-only} TBRU. Under these settings, the pure
encoder/decoder seq2seq model simply does not have the capacity to parse
newswire text with any degree of accuracy, but the TBRU-based approach is nearly
state-of-the-art {\em at the same exact computational cost}. 
As a point of comparison and an alternative to using input pointers, we also
implemented an attention mechanism within DRAGNN. We used the dot-product
formulation from \newcite{parikh2016decomposable}, where $\br(s_i)$ in the parser
takes in all of the {\em shift-only} TBRU's hidden states and $\mathbf{RNN}$
aggregates over them.

\paragraph{Utilizing parse representations improves summarization}

We evaluate our approach on the summarization task in Table \ref{tab:compress}.
We compare two single-task LSTM tagging baselines against two multi-task
approaches: an adaptation of \newcite{DBLP:journals/corr/LuongLSVK15} and the
stack-propagation idea of \newcite{zhang2016stack}. In both multi-task setups, we
use a right-to-left {\em shift-only} TBRU to encode the input, and connect it to
both our compositional {\em arc-standard} dependency parser and the
{\em Keep/Drop} summarization tagging model. 

In both setups we do not follow seq2seq, but utilize the $\textsc{Input}$
function to connect output decisions directly to input token
representations. However, in the stack-prop case, we use the $\textsc{Subtree}$
function to connect the tagging TBRU to the parser TBRU's phrase representations
directly (Figure \ref{fig:overview-multi-task}). We find that allowing the
compressor to directly use the parser's phrase representations significantly
improves the outcome of the multi-task learning setup. In both setups, we
pretrained the parsing model for 400K steps and tuned the subsequent ratio of
parser/tagger update steps using a development set.



\paragraph{Deep stacked bi-directional parsing}

Here we propose a continuous version of the bi-directional parsing model of
\newcite{attardi2009reverse}: first, the sentence is parsed in the left-to-right
order as usual; then a right-to-left transition system analyzes the sentence in
reverse order using addition features extracted from the left-to-right
parser. In our version, we connect the right-to-left parsing TBRU directly
to the phrase representations of the left-to-right parsing TBRU, again using the
$\textsc{Subtree}$ function. Our parser has the significant advantage that
the two directions of parsing can affect each other during training. During each
training step the right-to-left parser uses
representations obtained using
the {\em predictions} of the left-to-right parser. Thus, the
right-to-left parser can backpropagate error signals through the left-to-right parser
and reduce cascading errors caused by the pipeline.

Our final model uses 5 TBRU units. Inspired by \newcite{zhang2016stack}, a
left-to-right POS tagging TBRU provides the first layer of representations. 
Next, two {\em shift-only} TBRUs, one in each direction, provide representations
to the parsers. 
Finally, we connect the left-to-right parser to the right-to-left parser using
links defined via the $\textsc{Subtree}$ function. The result (Table
\ref{tab:deep-bidir}) is a state-of-the-art dependency parser, yielding the
highest published accuracy on the Treebank Union setup for both part of speech
tagging and parsing.

\section{Conclusions}

We presented a compact, modular framework for describing recurrent neural
architectures. We evaluated our dynamically structured model and found it
to be significantly more efficient and accurate than attention mechanisms for
dependency parsing and extractive sentence summarization in both single- and
multi-task setups. While we focused primarily on syntactic parsing, the framework
provides a general means of sharing representations between tasks. There remains
low-hanging fruit still to be explored: in particular, our approach can
be globally normalized with multiple hypotheses in the intermediate
structure. We also plan to push the limits of multi-task learning by combining
many different NLP tasks, such as translation,
summarization, tagging problems, and reasoning tasks, into a single model.




\bibliography{acl2017}
\bibliographystyle{acl_natbib}

\end{document}